%% file: arxiv_version.tex
\documentclass{article}
\usepackage{amsmath,graphicx,mlspconf}
\usepackage{xcolor}
\usepackage{enumitem}
\title{Geometry-Aware Graph Construction via\\
Adaptive Spectral Bandwidth Control\thanks{Accepted at IEEE MLSP 2026. \copyright 2026 IEEE. Personal use 
of this material is permitted. For any other use, permission must be 
obtained from the IEEE.}}
\name{Ecem Bozkurt, Antonio Ortega} 
\address{Department of Electrical and Computer Engineering\\ University of Southern California, Los Angeles, CA, USA \\ bozkurt@usc.edu, aortega@usc.edu}
\usepackage{cite}
\usepackage{amsmath,amssymb,amsfonts}
\usepackage{amsthm}
\usepackage{graphicx}
\usepackage{textcomp}
\usepackage{algorithm}
\usepackage{algorithmic}
\usepackage{xcolor}
\usepackage{booktabs}
\usepackage{multirow}
\usepackage{url}
\usepackage[hidelinks]{hyperref}
\usepackage{xcolor}

\def\BibTeX{{\rm B\kern-.05em{\sc i\kern-.025em b}\kern-.08em
    T\kern-.1667em\lower.7ex\hbox{E}\kern-.125emX}}

\newcommand{\reff}{r_{\mathrm{eff}}}

\input{macros}

\begin{document}
\maketitle

\begin{abstract}
Kernelized graph methods — spectral clustering, diffusion maps,
and sparse kernel-regression graphs — that use Gaussian kernels depend on the choice of Gaussian bandwidth $\sigma$, which governs the spectral character
of the local kernel operator.
When $\sigma$ is too small, the kernel overestimates local complexity
and treats each sample as an independent direction; when $\sigma$ is
too large, the kernel collapses multiple directions together, the
condition number diverges, and all geometric discrimination is lost.
We propose a choice of scale to make the spectral complexity of the kernel consistent with the intrinsic complexity
of the underlying manifold.
We propose a per-node bandwidth criterion that operationalizes this
principle by jointly matching the kernel's effective rank to the local intrinsic dimension estimated via minimum spanning tree, anchoring the search in the manifold-consistent log-log scaling regime.
We evaluate SSL embeddings from six encoders on CIFAR-100, showing that adaptive bandwidth consistently improves leave-one-out (LOO) classification and label propagation (LP) accuracy over fixed-bandwidth methods and competing adaptive methods.
\end{abstract}


\section{Introduction}
\label{sec:intro}

\noindent \textbf{Kernelized graphs as spectral operators.}
Modern self-supervised learning (SSL) encoders produce high-dimensional embeddings that serve as the input to downstream tasks such as classification, retrieval, clustering, and graph-based learning.
A large family of graph-based signal processing and learning methods
— spectral clustering~\cite{vonLuxburg2007}, Laplacian
eigenmaps~\cite{Belkin2003}, diffusion maps~\cite{CoifmanLafon2006},
and non-negative kernel (NNK) graphs~\cite{Shekkizhar2020} — build
a similarity graph by applying a Gaussian kernel to pairwise distances
and then operating on the resulting matrix.
In all of these methods, the bandwidth $\sigma$ is the main
hyperparameter.
We observe that without a careful choice of $\sigma$, the local characteristics in kernel space (e.g., its geometric properties estimated by the kernel matrix rank) can be completely different from
the local properties of the linear space (e.g., its local intrinsic dimension). 
The local kernel matrix can exhibit distinct spectral properties depending on the choice of $\sigma$, which impacts downstream tasks. 

\begin{figure}[t]
  \centering
    \includegraphics[width=0.88\linewidth]{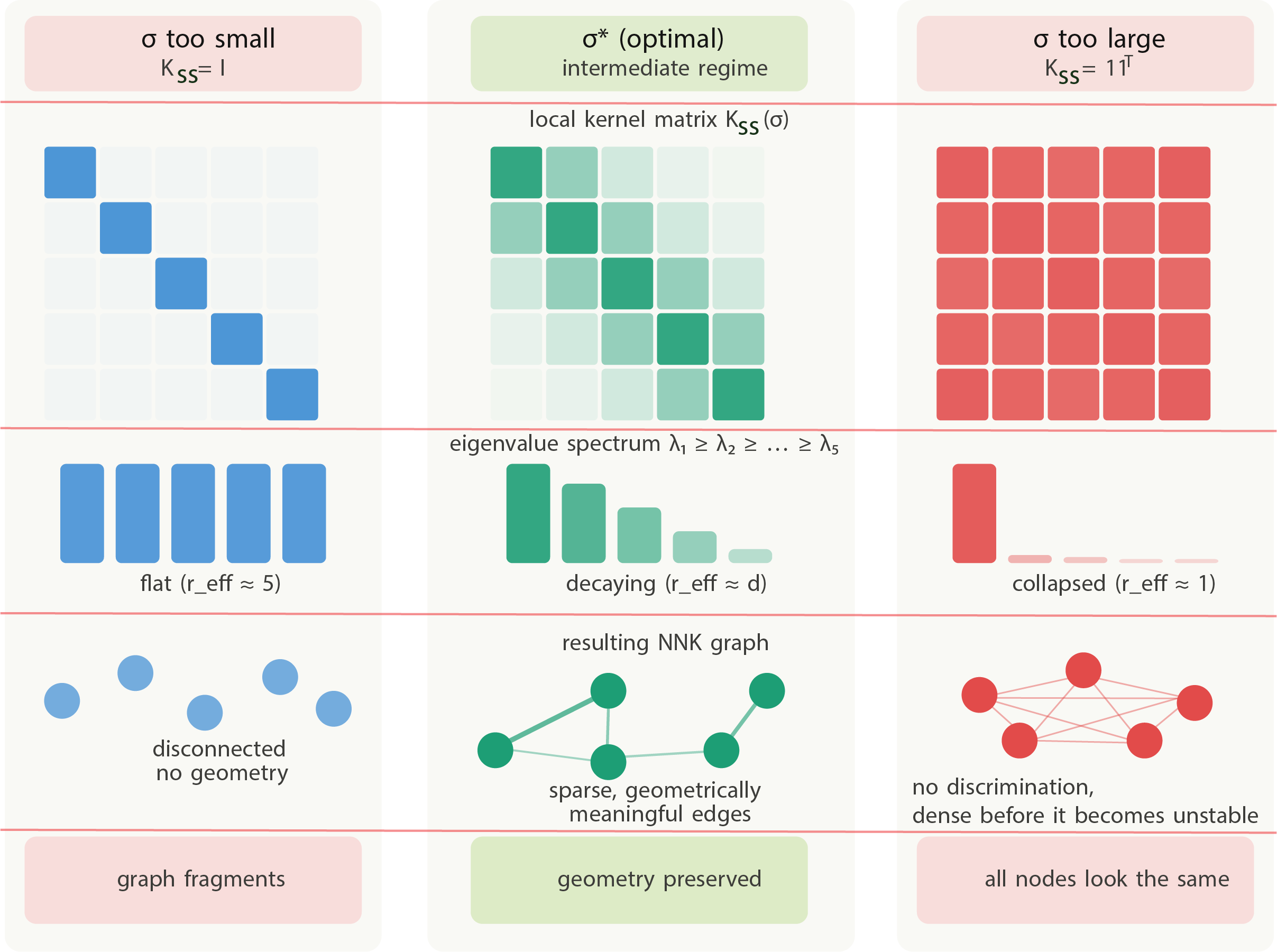}
  \caption{Spectral regimes of the local kernel operator $\Km_{SS}(\sigma)$.
    \emph{Left}: $\sigma$ too small — identity-like, flat spectrum,
    $\reff\approx|S|$, graph fragments, no geometric information.
    \emph{Center}: $\sigma^\star$ — decaying spectrum,
    $\reff\approx\hat{d}$, well-conditioned, geometrically informative.
    \emph{Right}: $\sigma$ too large — rank-one, $\kappa\to\infty$,
    all directions collapsed, discrimination lost.}
  \label{fig:regimes}
\end{figure}

\noindent\textbf{Two spectral failure modes.}
When $\sigma$ is too small, the Gaussian kernel approaches the
identity matrix: each data point appears orthogonal to every other,
effective rank is artificially inflated, the graph becomes
disconnected, and the spectral operator carries no useful structure.
When $\sigma$ is too large, the kernel collapses toward a rank-one
matrix: one eigenvalue dominates, the condition number $\kappa$
diverges, and the graph becomes a dense clique in which all geometric
discrimination is lost (Fig.~\ref{fig:regimes}).
Both regimes are useless for downstream tasks. Thus, we are interested in identifying where, between those two regimes, $\sigma$ should be chosen. 

\noindent \textbf{The missing principle.}
Existing bandwidth selection methods address parts of the problem
but miss a unifying principle (Table~\ref{tab:related}).
Distance-driven methods~\cite{Zelnik2004} anchor $\sigma$ to local
neighbor distances but do not consider the spectral character of the
resulting kernel or distance unreliability at higher dimensions.
Density normalization~\cite{CoifmanLafon2006} corrects sampling bias
but does not target spectral stability.
Log-log scaling methods~\cite{Singer2009,Lindenbaum2020} exploit the fact that, for data sampled from a low-dimensional manifold, the total kernel mass follows a predictable scaling law with respect to the bandwidth. 
The slope of this relationship can be used to estimate the manifold dimension and to identify scales at which the kernel provides a faithful approximation of the underlying geometry. 
Lindenbaum et al.~\cite{Lindenbaum2020} further select a bandwidth whose implied dimension matches an external intrinsic-dimension estimate. However, these approaches estimate a single global scale for the entire dataset, without adapting to local variations in density or geometry.
None of these methods selects $\sigma$ by asking whether 
the \emph{spectral complexity} — the effective number of independent
directions the kernel operator resolves, measured by the effective rank of the resulting kernel, is consistent
with the \emph{geometric complexity} of the local
neighborhood (the intrinsic dimensionality of the local data manifold). 

\noindent \textbf{A spectral-complexity principle.}
%
%
%
To avoid the spectral failure modes arising from poor choices of $\sigma$ (Fig.~\ref{fig:regimes}), we hypothesize that: \emph{a kernel operates at the correct scale when the complexity revealed through its spectrum is comparable to the intrinsic complexity of the local data manifold}.

\noindent \textbf{Connection to recent observations.}
Three recent results support this principle from different directions.
RankMe~\cite{Garrido2023} establishes that effective rank — the
entropy of the normalized eigenvalue distribution — quantifies the
number of active spectral degrees of freedom in a representation.
IDEST~\cite{Mordacq2026idest} shows that minimum spanning tree (MST)-based intrinsic dimension
tracks geometric structure in the high-dimensional, low-sample regime
where classical estimators break down.
T-REGS~\cite{Mordacq2025tregs} demonstrates that MST geometry is
directly regularizable and prevents spectral collapse.
Together, these results indicate that both spectral complexity and intrinsic dimension carry meaningful information about representation geometry. We therefore use effective rank as a measure of kernel-visible complexity and intrinsic dimension as a measure of manifold complexity, and seek bandwidths for which the two are locally consistent.

\par\noindent\textbf{Contributions} 
We propose  
(i) A per-node bandwidth criterion to select the scale at which the kernel's effective rank matches the local intrinsic dimension,
 placing the kernel operator in the informative scale, and 
(ii) Empirical evidence that $\sigma$ adaptation improves graph-based inference for both sparse and dense graph construction, and that dimMST is the appropriate complexity target under distance concentration.

\section{Related Work}
\label{sec:related}
\begin{table}[t]
\centering
\footnotesize
\caption{Bandwidth selection strategies.}
\label{tab:related}
\begin{tabular}{lll}
\toprule
Method & Strategy & Scale \\
\midrule
ZM~\cite{Zelnik2004}
& Neighbor distance
& Local \\

CL~\cite{CoifmanLafon2006}
& Density normalization
& Global \\

Singer~\cite{Singer2009}
& Log-log scaling
& Global \\

Lindenbaum~\cite{Lindenbaum2020}
& Dimension matching
& Global \\

\textbf{Ours}
& Effective rank + dimMST +log-log scaling
& Local \\
\bottomrule
\end{tabular}
\end{table}

\par\noindent
\textbf{Bandwidth selection for kernel graph operators} Closest to our work is 
Lindenbaum et al.~\cite{Lindenbaum2020}, 
where 
a global bandwidth and a feature-scaling matrix are selected to match a kernel-implied dimension to an external estimate.
Our criterion shares the dimension-matching spirit but differs in that (i) it is applied \emph{per node}, (ii) it uses effective
rank rather than kernel-implied dimension as the spectral measure,
and (iii) it estimates intrinsic dimension with dimMST~\cite{Mordacq2026idest}, which is robust to the
distance concentration that affects methods that operate on high-dimensional $\ell_2$-normalized embeddings.

\par\noindent
\textbf{Non-negative kernel (NNK) graphs}
NNK graphs~\cite{Shekkizhar2020} remove redundant neighbors while preserving local structure through a non-negative quadratic program in the kernel domain,  which can be interpreted as a geometric condition in linear space. 
In these methods, which have been used to study SSL and LLM
geometry~\cite{Cosentino2022,Balestriero2024,Hurtado2022},  $\sigma$ is treated as an external hyperparameter. In contrast, we focus on selecting the bandwidth itself, leaving the NNK optimization unchanged. 
%
While we test our $\sigma$ selection criterion to construct NNK graphs, the underlying idea is more general: selecting $\sigma$ adjusts the relative
importance of neighbors through the kernel weights, and is therefore
a form of neighborhood selection if the
neighborhood is defined explicitly (as in NNK) or implicitly through
weight decay (as in $k$-NN).

\par\noindent
\textbf{Effective rank and intrinsic dimension.}
We use effective rank~\cite{RoyVetterli2007} as the spectral complexity measure: it is continuous, entropy-based, and
captures how uniformly information is distributed across all
eigenvalue directions — unlike the participation ratio, which relies
only on second-order statistics, or raw rank, which is sensitive
to small eigenvalues.
Its recent adoption in RankMe~\cite{Garrido2023} further confirms that effective rank is a reliable indicator of how many meaningful directions a
kernel operator resolves.
Intrinsic dimension of the embedding, on the other hand,  indicates how many geometrically relevant directions exist in linear space. We use MST-based intrinsic dimension~\cite{Mordacq2026idest}, which 
is provably consistent under weaker assumptions than Levina--Bickel~\cite{LevinaBickel2005}
or TwoNN~\cite{Facco2017}, and remains informative under distance
concentration.

\section{Method}
\label{sec:method}

\subsection{Spectral characterization of the bandwidth problem}

For node $\xv_j$, with candidate neighbors $S_j$ and $\xv_k \in S_j$, define the local Gaussian
kernel matrix $\Km_{SS}(\sigma)$ with entries
$\Km_{jk}=\exp(-\|\xv_j-\xv_k\|^2/(2\sigma^2))$.
The intrinsic dimension $\hat{d}$ characterizes how many independent
directions exist in the local data manifold, working directly with the $\xv_i$, while the effective rank $\reff(\Km_{SS}(\sigma))$ characterizes how many independent directions the kernel operator actually resolves at a given $\sigma$. 
%

When these two quantities disagree, the kernel is misspecified (see Fig.~\ref{fig:regimes}):
if $\reff < \hat{d}$, the kernel collapses genuinely distinct manifold
directions into a single component, neighbors that lie in different directions become indistinguishable, and the graph loses
discriminative power.
If $\reff > \hat{d}$, the kernel resolves more directions than the
manifold has, treating noise and ambient dimensions as genuine
structure, causing the graph to fragment along spurious directions.
In both cases, the mismatch corrupts the graph's geometric fidelity.

\noindent\textbf{Target regime.}
We seek an operating point where the kernel sees a complexity consistent with what the manifold
provides, i.e., 
$\reff(K_{SS}(\sigma^\star))\approx\hat{d}$, 
where $\Km_{SS}(\sigma)$ resolves approximately
$\hat{d}_i$ independent directions, matching the intrinsic
dimensionality of the local neighborhood (see Fig.~\ref{fig:regimes}(center)).
In this regime, the condition number  $\kappa$ is moderate, the graph is well-conditioned,
and the spectral operator captures the local geometry of the manifold.
Next, we introduce the local metrics we use to optimize $\sigma$.  



\subsection{Local neighborhood metrics}

\par\noindent\textbf{Effective rank} We use the effective rank~\cite{RoyVetterli2007} of $\Km_{SS}(\sigma)$: 
\begin{equation}
\reff(\Km) = \exp\!\Bigl(-\textstyle\sum_j p_j\log p_j\Bigr), 
\label{eq:erank}
\end{equation}
where $p_j=\lambda_j/\sum_k\lambda_k$. 
From a signal-processing perspective, $\reff$ quantifies the number
of \textbf{active spectral degrees of freedom represented by the kernel
operator}: $\reff=1$ indicates a rank-one operator (all energy in one
direction); $\reff=|S|$ indicates a uniform operator (no dominant
direction).
The target is $\reff\approx\hat{d}_i$: the kernel should resolve
as many independent directions as the local manifold has.

\par\noindent\textbf{Intrinsic dimension and manifold complexity.}
We estimate the local intrinsic dimension $\hat{d}_i$ using the minimum spanning tree
(MST)-based estimator (dimMST)~\cite{Mordacq2026idest} on the
$k_{\mathrm{cand}}$ neighborhood.
Nearest-neighbor estimators such as
Levina--Bickel~\cite{LevinaBickel2005} and TwoNN~\cite{Facco2017} rely on ratios of inter-point distances; on $\ell_2$-normalized embeddings, these ratios lose contrast, and the estimators become unstable at higher dimensions, as the distances concentrate.
The dimMST estimator depends on the growth
rate of the spanning tree rather than on pairwise distance contrasts.
IDEST~\cite{Mordacq2026idest} shows that dimMST tracks SSL
representation quality in the $n \approx d$ regime where TwoNN and MLE fail. 
Our results (Fig.~\ref{fig:erank_dim}) confirm this
empirically: dimMST aligns with $\reff$ at $\sigma^\star$
substantially better than Levina--Bickel across all six encoders.

\par\noindent\textbf{Manifold scaling consistency and log-log slope .}
Following Singer~\cite{Singer2009} and Lindenbaum~\cite{Lindenbaum2020},
on a $\hat{d}$-dimensional manifold the total kernel energy
$L_i(\sigma)=\sum_{j,k\in S_i}\Km_{jk}(\sigma)$ satisfies
$\log L_i\approx (\hat{d}/2)\log\sigma + C$ in the manifold-consistent
regime.

Let $\ell_i(\sigma) = d\log L_i(\sigma)/d\log\sigma$ denote the
log-log slope of the total kernel energy at node~$i$, and
$\ell_{\max} = \max_{\sigma\in\mathcal{G}_i}\ell_i(\sigma)$ the
peak slope over the local bandwidth grid.
We compute $\ell_{\max}=\max_\sigma\ell_i(\sigma)$ to locate the peak of the consistent scaling region. It penalizes bandwidths that fall outside this linear regime.
\subsection{Bandwidth optimization criterion}

To achieve our desired bandwidth target, we combine the previously introduced metrics to define an optimality criterion:
\begin{equation}
J_i(\sigma)
= \underbrace{\frac{|\reff(K_{SS}(\sigma))-\hat{d}_i|}{\hat{d}_i}}_{\text{spectral}\leftrightarrow\text{manifold complexity}}
  +
  \underbrace{\frac{\ell_{\max}-\ell_i(\sigma)}{\ell_{\max}}}_{\text{scaling consistency}},
\label{eq:score}
\end{equation}
Both terms are dimensionless and expressed as fractional quantities, placing them on a common scale without requiring an additional balancing coefficient.
The first term drives the effective rank toward the local intrinsic
dimension, penalizing both the identity limit ($\reff\gg\hat{d}$)
and the rank-one limit ($\reff\ll\hat{d}$).
The second term selects the bandwidth at which the kernel's energy
scaling is most consistent with manifold geometry; note that it
has no knowledge of the local intrinsic dimension — it defines a
valid operating region, but cannot distinguish between a kernel
that sees too few directions and one that sees too many within
that region.
The effective-rank term resolves this ambiguity.

Our target bandwidth is then 
\begin{equation}
\sigma_i^\star
= \arg\min_{\sigma\in\mathcal{G}_i}
  J_i(\sigma),
\end{equation}

which we find via a local log-spaced grid search, with 
\[\mathcal{G}_i = \mathrm{logspace}(0.05\,d_{i,k_{\mathrm{mle}}},\,
3.0\,d_{i,k_{\mathrm{cand}}})\] where, in our experiments, we choose $|\mathcal{G}|=12$. 


The per-node cost of the bandwidth search is
$O(|\mathcal{G}| \cdot k_{\mathrm{cand}}^2)$ for kernel matrix
construction and energy summation, plus
$O(|\mathcal{G}| \cdot k_{\mathrm{cand}}^2)$ for the eigenvalue
computation, giving $O(|\mathcal{G}| \cdot k_{\mathrm{cand}}^2)$
overall since $k_{\mathrm{cand}} \ll n$.
Given the number of grid points,$|\mathcal{G}|$, and $k_{\mathrm{cand}}$, this is a
constant-time operation per node; the total graph construction cost
is $O(n \cdot k_{\mathrm{cand}}^2)$, dominated in practice by the $k$-NN search.

\subsection{Weight sharpening: a secondary refinement}

As a secondary refinement, we adjust the edge weights without changing the neighborhood support. The intuition is that dense regions, where neighbors
are close and distances are small, should place more weight on their strongest neighbors, whereas sparse regions should distribute weight more evenly to avoid relying on a single connection.

The selected bandwidth $\sigma_i^\star$ already reflects local density: small bandwidths typically occur in dense regions and large bandwidths in sparse ones. We therefore define a per-node exponent
\begin{equation}
p_i=\mathrm{clip}
\left((\sigma_i^\star/\sigma_g)^{-1},0.2,2.0
\right),
\label{eq:sharpen}
\end{equation}
where $\sigma_g=\mathrm{median}_j,\sigma_j^\star$ is the global reference bandwidth. The clipping operation prevents extreme weight concentration or flattening. The edge weights are then updated as
\[
w_j\leftarrow\frac{w_j^{p_i}}{\sum_k w_k^{p_i}}.
\]
When $p_i>1$, larger weights become more dominant, concentrating support on the strongest neighbors. When $p_i<1$, the weights become more uniform, distributing support across multiple neighbors. This only redistributes weight among existing edges and does not modify the graph topology.


\section{Experiments}
\label{sec:experiments}

\subsection{Setup}

The proposed bandwidth-selection criterion is applied prior to the 
graph-construction procedure and can therefore be used for both
dense and sparse graph methods.
For NNK, we solve the standard non-negative quadratic program
at scale $\sigma_i^\star$.
For dense graph baselines, we use Gaussian-weighted $k$-NN with
the same bandwidth.

\noindent\textbf{Ablation design}
The following chain of baselines isolates the contribution of each
component:
\[
\underbrace{\text{kNN}}_{\text{no }\sigma}
\to
\underbrace{\text{kNN-}\sigma}_{\sigma^\star}
\to
\underbrace{\text{kNN-}\sigma\text{-}\alpha}_{\sigma^\star+\text{sharpen}}
\to
\underbrace{\text{NNK-fixed}}_{\text{global }\sigma}
\to
\underbrace{\text{NNK-}\sigma}_{\sigma^\star+\text{sparse}}
\]
kNN-$\sigma$ uses $\sigma_i^\star$ as a per-node Gaussian bandwidth
over $k_{\mathrm{cand}}$ neighbors without sparsification.
Each step adds exactly one component, enabling clean attribution in
the LOO results.

\noindent\textbf{Datasets and embeddings.} Tests use six SSL encoders on 
CIFAR-100~\cite{Krizhevsky2009}: SimCLR~\cite{Chen2020simclr},
MoCo\,v2~\cite{Chen2020mocov2}, BYOL~\cite{Grill2020},
Barlow Twins~\cite{Zbontar2021}, VICReg~\cite{Bardes2022},
DINO~\cite{Caron2021}, from \cite{Turrisi2022}.
ResNet-50~\cite{He2016}: $d=512$; ViT-S/16~\cite{Dosovitskiy2021}
(DINO): $d=384$. All embeddings $\ell_2$-normalized, 3{,}000 samples (30 per class).

\noindent \textbf{Hyperparameters}  
(Table~\ref{tab:hyperparams}). We chose a candidate pool size of \( k_{\mathrm{cand}} = 30 \) to balance neighborhood coverage and computational cost, as larger sizes improve dimMST accuracy but increase the per-node eigenvalue cost at \( O(k^3) \).
Using \( |\mathcal{G}| = 12 \) provides sufficient log spacing to identify both the slope peak and the crossing of \( \reff = \hat{d} \), while \( k_{\mathrm{mle}} = 10 \) anchors the lower bound to the MLE-implied scale.
To reduce variance in the per-node dimension estimate, we perform \( n_{\mathrm{rep}} = 5 \) replications of dimMST, which adds a fixed number of MST computations per node.
In the LP method~\cite{Zhou2004}, we set \( \alpha = 0.85 \), perform 200 iterations with early stopping, and adjust the labeled rows of the transition matrix to zero to ensure they act purely as sources without absorbing beliefs from neighbors.
In our LOO strategy, graphs are completely rebuilt at each sparsity level. 
All results are averaged over five independent stratified trials.
\begin{table}[h]
\centering
\footnotesize
\caption{Hyperparameter summary.}
\label{tab:hyperparams}
\begin{tabular}{ll}
\toprule
Parameter & Value \\
\midrule
Candidate pool $k_{\mathrm{cand}}$ & 30 \\
MLE anchor $k_{\mathrm{mle}}$ & 10 \\
Min.\ support $k_{\mathrm{min}}$ & 6 \\
Bandwidth grid $|\mathcal{G}|$ & 12 \\
dimMST replications $n_{\mathrm{rep}}$ & 5 \\
LP diffusion $\alpha$ & 0.85 \\
LP iterations & 200 \\
Trials & 5 \\
\bottomrule
\end{tabular}
\end{table}

\subsection{Local neighborhood evaluation}
\label{sec:q1}

\begin{figure}[ht]
  \centering
  \includegraphics[width=0.94\linewidth]
  {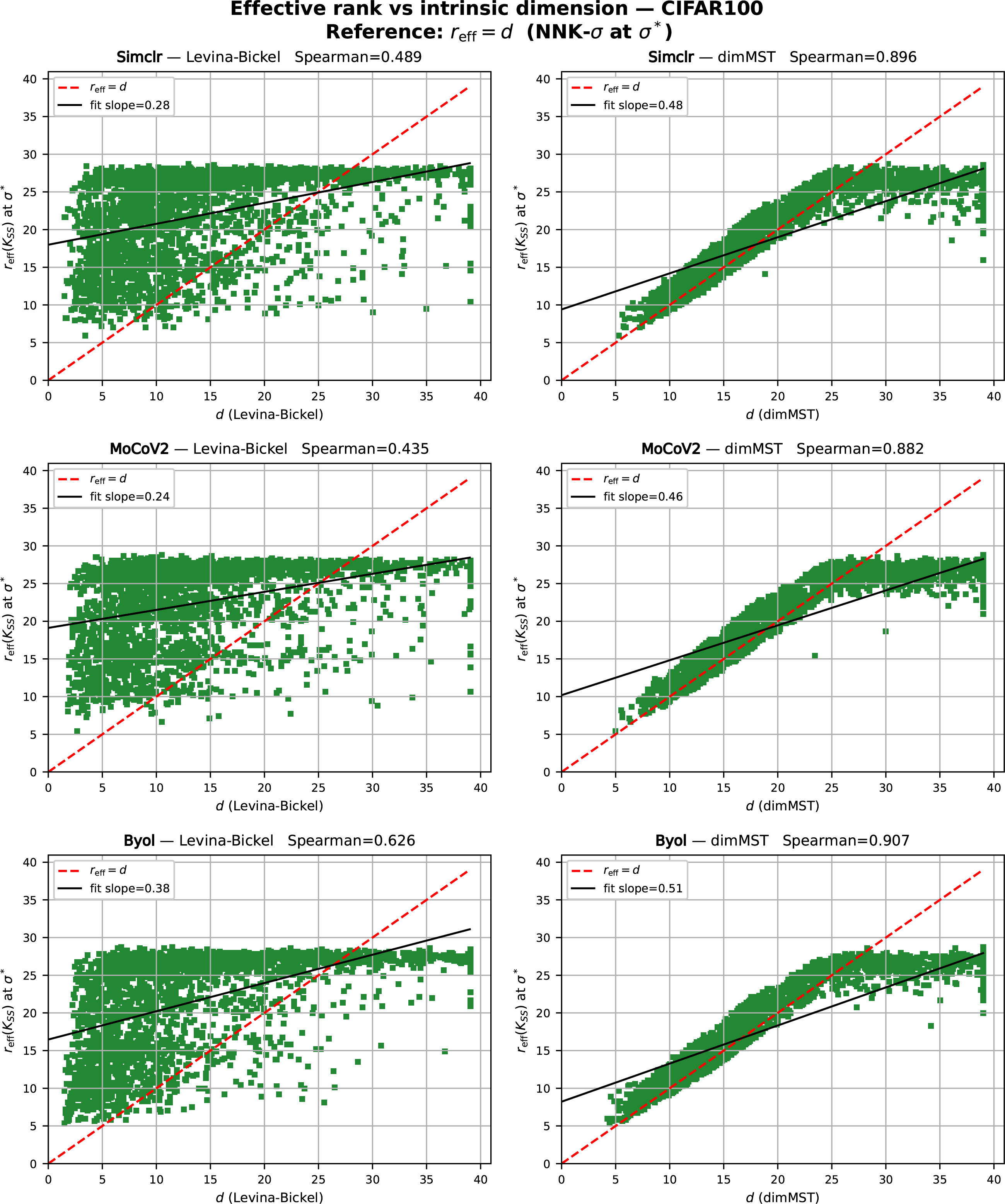}
  \caption{$\reff(K_{SS})$ at $\sigma^\star$ vs.\ per-node intrinsic
    dimension for 3 encoders (rows) for the 
     LB estimator (Left) and dimMST (Right). 
    dimMST aligns substantially better, confirming it as a better complexity target under distance concentration.}
  \label{fig:erank_dim}
\end{figure}

We test whether a $\sigma^\star$ that optimizes \eqref{eq:score}
produces local kernel matrices whose effective rank is aligned with
the local intrinsic dimension.
Fig.~\ref{fig:erank_dim} plots $\reff(K_{SS})$ at $\sigma^\star$
against the per-node dimension estimate, comparing dimMST to Levina--Bickel (LB) for three encoders.

Spearman correlations between $\reff$ and dimMST range from 0.88 to 0.91 on CIFAR-100 — consistently
and substantially higher than the correlations with LB (0.43--0.63).
This confirms that \textbf{dimMST is the appropriate target}:
LB intrinsic dimension estimators are degraded by distance concentration on the $\ell_2$ sphere, while the MST-based intrinsic dimension estimation remains informative.

The \textbf{fit slopes} (0.3--0.5) are below the ideal $\reff=\hat{d}$ line, due 
to finite-sample saturation of dimMST for  $k_{\mathrm{cand}}=30$.
However, as shown in downstream task experiments, even without an exact matching, consistent \emph{rank ordering} of $\hat{d}_i$ across nodes leads to improved performance. 


\subsection{Leave-One-Out (LOO) classification}
\label{sec:q2}
\begin{figure}[ht]
  \centering
  \includegraphics[width=1\linewidth]
  {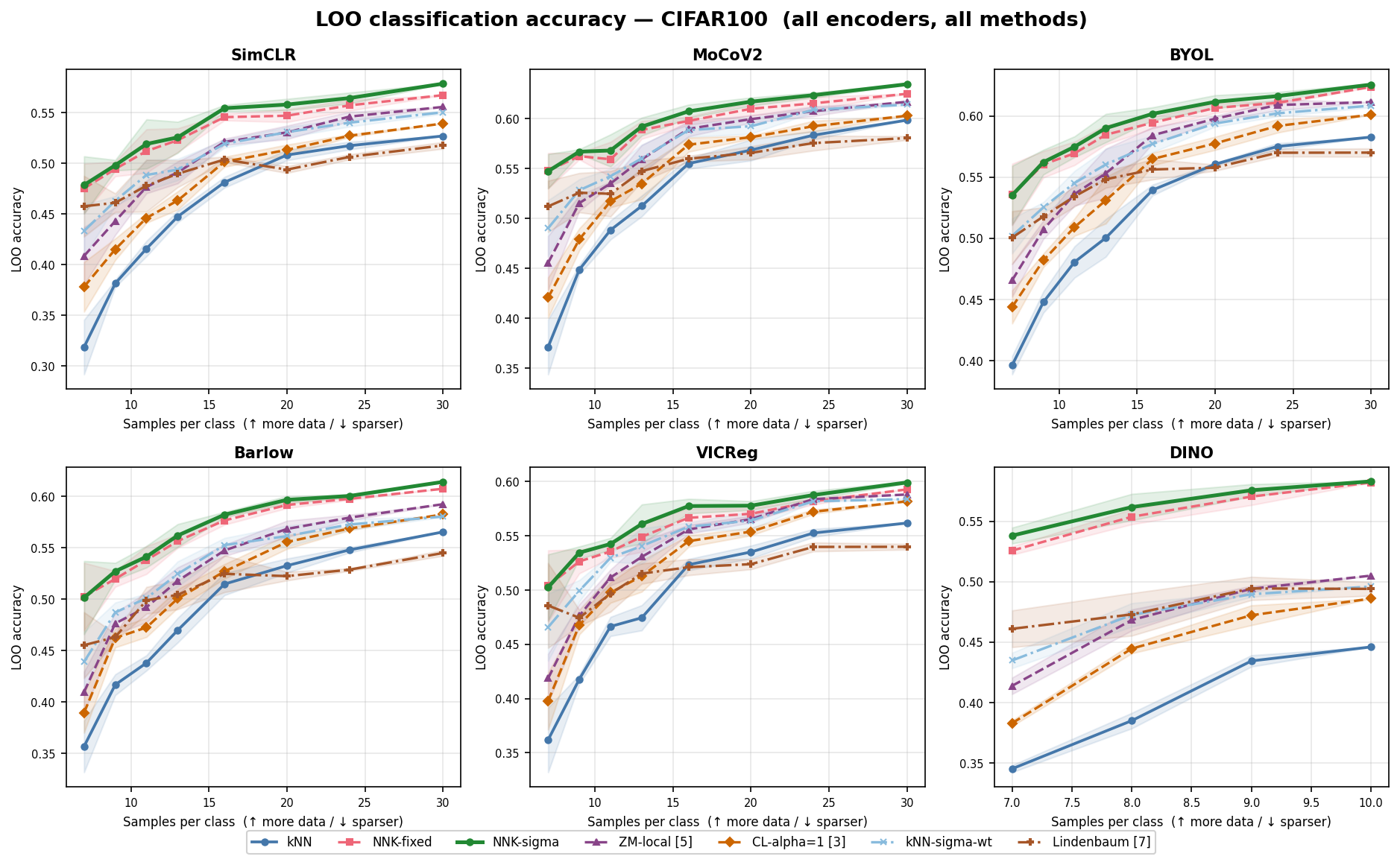}
  \caption{LOO accuracy vs.\ samples per class,
    CIFAR-100, six encoders.
    NNK-$\sigma$ (green) leads consistently across all encoders. Sigma adaptation improves both dense graphs (kNN $\to$ kNN-$\sigma$-$\alpha$) and sparse graphs (NNK-fixed $\to$ NNK-$\sigma$).}
  \label{fig:loo}
\end{figure}

LOO directly measures per-node graph quality (Fig.~\ref{fig:loo}): each node is predicted by a weighted majority vote over its graph neighbors, testing whether the graph's local spectral structure supports accurate inference.
For \textit{dense} graphs, kNN-$\sigma$-$\alpha$ consistently outperforms
plain kNN across all encoders, with the largest gaps at low
$n_{\mathrm{pc}}$ where bandwidth choice is most critical.
For \textit{sparse} graphs, NNK-$\sigma$ outperforms NNK-fixed in all cases.
The gain from NNK-fixed to NNK-$\sigma$ shows that geometric sparsification at the correct $\sigma^\star$ retains only the
geometrically relevant edges.
The ordering kNN $<$ kNN-$\sigma$-$\alpha$ $<$ NNK-fixed $<$ NNK-$\sigma$ holds, sigma adaptation helps both dense and sparse graphs, and the sparse solve provides a consistent further gain on top of sigma adaptation. NNK-$\sigma$ leads
across all six encoders at all label fractions.

\begin{table}[h]
\centering\footnotesize
\caption{LOO accuracy vs.\ dimension target scale $\gamma$.}
\label{tab:dscale}
\begin{tabular}{ccccc}
\toprule
$\gamma$ & LOO & $\mathrm{med}\,\reff$ & $\mathrm{med}\,\hat{d}$ \\
\midrule
0.50 & 0.607 & 14.7 & 13.3 \\
0.75 & 0.607 & 21.0 & 20.0 \\
1.00 & 0.611 & 25.7 & 26.7 \\
1.25 & 0.611 & 26.1 & 33.3 \\
1.50 & \textbf{0.612} & 26.3 & 40.0 \\
2.00 & 0.611 & 26.4 & 53.3 \\
\bottomrule
\end{tabular}
\end{table}

\noindent The effective-rank term targets $\reff \approx \gamma\hat{d}$, where $\gamma=1$ scales the dimMST estimate.
Table~\ref{tab:dscale} reports LOO accuracy and median $\reff$ on CIFAR100-MoCoV2 (npc=20) as $\gamma$ varies.
When $\gamma < 0.75$, the criterion forces $\reff$ below the manifold dimension, selecting a bandwidth that is too small: the kernel sees fewer directions than the data has, and LOO accuracy drops.
When $\gamma \geq 1.0$, the criterion is robust once the target is at or above the true dimension.
At $\gamma = 2.0$ LOO accuracy drops again.
This confirms the role of the effective-rank term: it prevents the bandwidth from collapsing to the identity limit. 

\subsection{Label Propagation (LP) accuracy}
\label{sec:q3}

\begin{figure}[ht]
  \centering
  \includegraphics[width=1\linewidth]{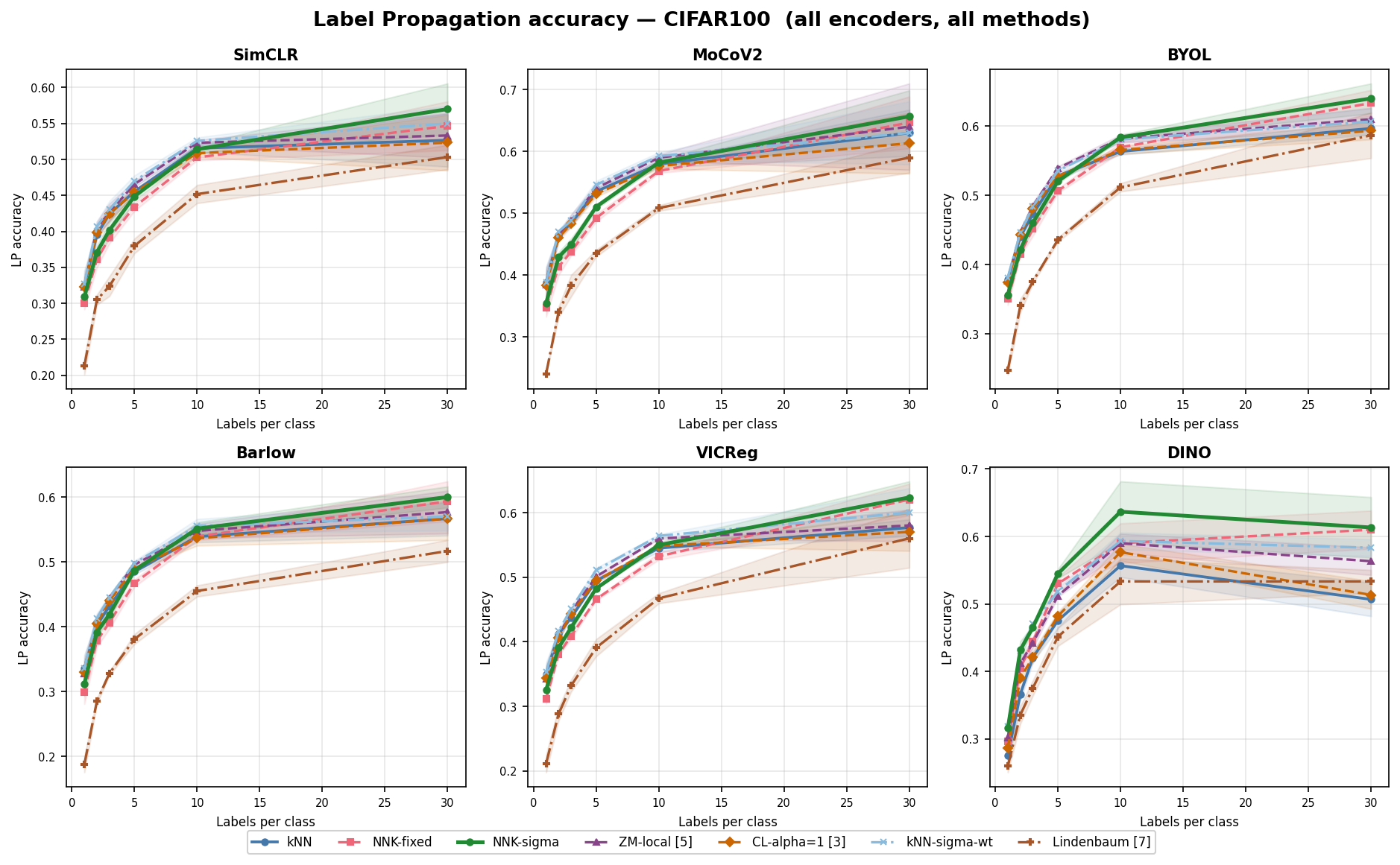}
  \caption{LP accuracy vs.\ labels per class, six encoders.
    NNK-$\sigma$ (green) leads on CIFAR-100 across all encoders, with gains over NNK-fixed most pronounced at $\geq$10 labels per class.}
  \label{fig:lp}
\end{figure}

LP \cite{ZhuGhahramaniLafferty2003,Zhou2004}  tests whether the graph's spectral structure supports
semi-supervised inference over the full graph (Fig.~\ref{fig:lp}).
On CIFAR-100 (100 classes, harder propagation), 
NNK-$\sigma$ leads
across all six encoders at all label fractions.
Gains are most pronounced at $\geq$10 labels/class, where graph
quality becomes the binding constraint rather than label sparsity.
NNK-$\sigma$+$\alpha$ provides a further improvement over NNK-$\sigma$
at higher label fractions, suggesting that weight sharpening
complements bandwidth selection when sufficient labels are
available to propagate beyond immediate neighbors.

\section{Conclusion}
\label{sec:conclusion}

We have proposed 
a per-node bandwidth criterion that matches the kernel's active
spectral degrees of freedom to the local intrinsic dimension of the data manifold consistently produces better-conditioned graph operators
and improves downstream performance.
%
Our results show that $\sigma$ adaptation helps regardless of whether the graph is
sparse (NNK) or dense ($k$-NN): the spectral alignment at $\sigma^\star$
improves graph conditioning and inference accuracy for all encoders. 
Additionally, geometric sparsification at the selected $\sigma^\star$
provides a consistent additional gain over sigma-adapted $k$-NN,
confirming that bandwidth selection and sparse graph construction
are complementary.
%
The criterion~\eqref{eq:score} is not specific to NNK graphs.
It can be applied before any Gaussian-kernel graph construction by
running the per-node grid search and passing $\sigma_i^\star$ to the downstream method.
The key principle — that kernel effective rank should be consistent with local intrinsic dimension — connects spectral graph signal
processing to the geometry of the underlying data manifold.

\small
\bibliographystyle{IEEEbib}
\bibliography{refs} 


\end{document}

%% file: macros.tex
\long\def\comment#1{}

\newfont{\bbb}{msbm10 scaled 700}

\newfont{\bb}{msbm10 scaled 1000}

\newcommand{\xv}{{\bf x}}

\newcommand{\Km}{{\bf K}}

\renewcommand{\arg}{{\hbox{arg}}}